# Dynamical Systems based Obstacle Avoidance with Workspace Constraint for Manipulators

Dake Zheng[1, 2], Xinyu Wu[1], and Jianxin Pang[2]

(dake.zheng@ubtrobot.com, xy.wu@siat.ac.cn, and walton@ubtrobot.com)

*Abstract*— In this paper, based on Dynamical Systems (DS), we present an obstacle avoidance method that take into account workspace constraint for serial manipulators. Two modulation matrices that consider the effect of an obstacle and the workspace of a manipulator are determined when the obstacle does not intersect the workspace boundary and when the obstacle intersects the workspace boundary respectively. Using the modulation matrices, an original DS is deformed. The proposed approach can ensure that the trajectory of the manipulator computed according to the deformed DS neither penetrate the obstacle nor go out of the workspace. We validate the effectiveness of the approach in the simulations and experiments on the left arm of the UBTECH humanoid robot.

## I. INTRODUCTION

It is well known that each serial manipulator has a corresponding workspace, therefore the given trajectory of its end-effector must be within the workspace. In practice, manipulators often work with other agents and objects, e.g. bottles, human workers and other manipulators. The agents and objects can be referred to as obstacles. When an obstacle appears on the given trajectory, the manipulator can not track the original trajectory anymore. A new trajectory needs to be computed to avoid the obstacle. In addition, the newly computed trajectory must be in the workspace.

Obstacle avoidance has been studied for a long time in robotics, and many methods have been proposed. Generally, those methods can be divided into two categories, i.e. the global approaches and the local approaches. Local approaches such as the vector field histogram [1] and the curvature-velocity method [2] can avoid the obstacles rapidly. However, those methods are usually locally optimal, they may fail to get a feasible path.

Global approaches such as the probabilistic roadmaps method [3] and the rapidly exploring random tree method [4] avoid obstacles by using path planning algorithms. Those methods can always find a collision-free path even in very complex scenarios. Although it is possible to parallelise the algorithms, the computational costs of the global path searches for those methods are still very heavy. Therefore, those methods cannot be applied to real-time obstacle avoidance [5].

In order to achieve real-time obstacle avoidance, several methods have been proposed. In the presence of obstacles, the elastic band approach [6, 7] deforms the original path by applying repulsive forces to get a new collision-free path. A reactive motion planning approach is proposed in [8], during the execution of a task, this approach avoids obstacles by re-planning and deforming the original path.

Khatib [9] proposes the artificial potential field method. Based on this method, Park et al. [10] propose the dynamic potential field methods, Iossifidis and Schöner [11] propose the attractor dynamics approach, Sprunk et al. [12] propose a kinodynamic trajectory generation method, etc. The artificial potential field method models each obstacle with a repulsive force to avoid collision between the robot and the obstacle. The repulsive force should be well defined to avoid local minima. To overcome the limitations of the potential field methods, the harmonic potential methods [13, 14] are proposed and widely used [15]. This approach is inspired by the description of the dynamics of fluids around impenetrable obstacles.

Similar to the harmonic potential functions method, Khansari Zadeh et al. [16] propose the dynamical systems (DS) based method recently. A trajectory can be computed according to an original DS. In the presence of obstacles, the original DS is deformed by a modulation matrix of the obstacles, then, a new trajectory that can avoid the obstacles is computed according the deformed DS. Compared to the harmonic potential functions method, the DS based method does not have to follow harmonic functions so it can be applied more widely. Huber et al. [17] extend out the DS based approach and propose an approach to avoid multiple concave obstacles. However, the approach is proposed under the assumption that the original DS is linear, therefore, the application of the approach is limited.

Since manipulators usually require real-time obstacle avoidance, hence the DS based method is a good choice for the obstacle avoidance of manipulators. As stated above, a manipulator must work in its workspace, however, few current obstacle avoidance methods take into account the workspace constraint. Therefore, the trajectories computed by the current methods may tend to go out of the workspace, then the manipulators will not work properly. For a serial manipulator, in order to get obstacle avoidance in the workspace of the manipulator, we extend out the work in [16] and propose a DS based approach.

## II. PROBLEM FORMULATION

Fig.1(a) illustrates the left arm of the UBTECH humanoid robot tracks a trajectory computed from a modulated DS proposed in [16]. The DS is obtained by applying an obstacle modulation matrix to an original DS, and is used to avoid the obstacle. Suppose the original trajectory computed from the original DS is in the workspace of the arm. The blue point cloud represents the workspace. When the obstacle is placed in the current position, the trajectory computed from the modulated DS goes out of the workspace, although it avoids the obstacle. Since part of the trajectory is out of the workspace, hence, the arm will fail to track the trajectory and will cause some damage.

[1] Shenzhen Institutes of Advanced Technology, Chinese Academy of Sciences, Shenzhen 518055 P.R.China.

[2] UBTECH Robotics, Inc., Shenzhen 518055 P.R.China.

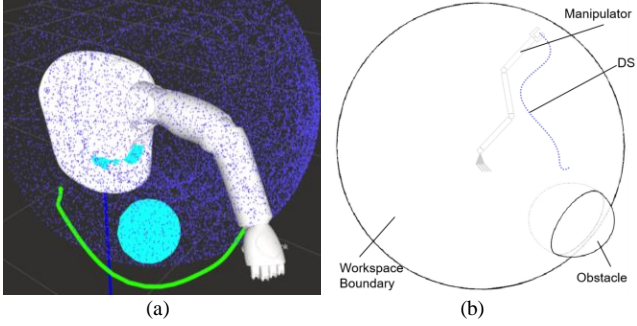

Figure 1. A manipulator tracking a DS based trajectory.

For the above problem, a method will be proposed to ensure that the DS based trajectory neither go out of the workspace nor penetrate the obstacle. Fig.1(b) shows a simplified model of the problem. For simplicity, a spherical workspace and a convex obstacle are considered in this paper.

## III. OBSTACLE-AVOIDANCE ALGORITHM

Consider a serial manipulator in Fig.1(b), we denote a state variable $\xi$ as the translational position of the end-effector of the manipulator. A DS based trajectory for the end-effector can be computed according to:

$$\dot{\xi} = f(\xi), \quad f : \Re^3 \mapsto \Re^3 \tag{1}$$

where $f(\cdot)$ is a continuous function. Given a start position $\{\xi\}_0$, the trajectory of the end-effector can be computed along time according to:

$$\{\xi\}_t = \{\xi\}_{t-1} + f(\{\xi\}_{t-1})\Delta t \tag{2}$$

where $t$ is a positive integer and $\Delta t$ is the integration time step.

In this paper, we suppose the original DS given by Eq.(1) is a three-dimensional globally asymptotically stable and converges to a point $\xi^*$, i.e. $f(\xi^*) = 0$, in the workspace.

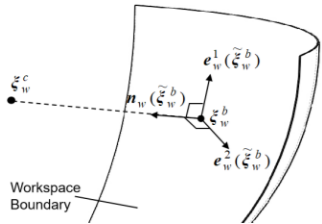

Figure 2. Illustration of the workspace boundary.

### A. Convex Obstacle and Workspace

As shown in Fig.1(b), the workspace boundary and the obstacle surface are convex. $\xi_o^c$ and $\xi_w^c$ represent the obstacle center and the workspace boundary center, respectively. Define $\tilde{\xi}_o = \xi - \xi_o^c$ and $\tilde{\xi}_w = \xi - \xi_w^c$, then, the obstacle surface and the workspace boundary can be described by the two three-dimensional ellipsoids, respectively, as follows:

$$\Gamma_o(\tilde{\xi}_o) : \sum_{i=1}^{3}((\tilde{\xi}_o)_i / a_i^o)^{2p_o} = 1 \tag{3}$$

$$\Gamma_w(\tilde{\xi}_w) : \sum_{i=1}^{3}((\tilde{\xi}_w)_i / a_i^w)^{2p_w} = 1 \tag{4}$$

where functions $\Gamma_o(\tilde{\xi}_o)$ and $\Gamma_w(\tilde{\xi}_w)$ are continuous distance functions and have first order partial derivatives and increases monotonically with $\|\tilde{\xi}_o\|$ and $\|\tilde{\xi}_w\|$, respectively. $(\cdot)_i$ represents the $i$-th element value of a vector $(\cdot)$. $a_i^o$ and $a_i^w$ are the $i$-th axis lengths of the obstacle and the workspace boundary, respectively. $p_o$ and $p_w$ are positive integers.

As in [16], the functions $\Gamma_o(\tilde{\xi}_o)$ and $\Gamma_w(\tilde{\xi}_w)$ can divide the obstacle and the workspace into exterior, boundary and interior regions, respectively, according to:

$$\begin{cases} \chi_o^e = \{\xi \in \Re^3 : \Gamma_o(\tilde{\xi}_o) > 1\} \\ \chi_o^b = \{\xi \in \Re^3 : \Gamma_o(\tilde{\xi}_o) = 1\} \\ \chi_o^i = \{\xi \in \Re^3 : \Gamma_o(\tilde{\xi}_o) < 1\} \end{cases}, \begin{cases} \chi_w^e = \{\xi \in \Re^3 : \Gamma_w(\tilde{\xi}_w) > 1\} \\ \chi_w^b = \{\xi \in \Re^3 : \Gamma_w(\tilde{\xi}_w) = 1\} \\ \chi_w^i = \{\xi \in \Re^3 : \Gamma_w(\tilde{\xi}_w) < 1\} \end{cases} \tag{5}$$

where $\chi_o^e$, $\chi_o^b$ and $\chi_o^i$ are points in exterior, boundary and interior regions of the obstacle, respectively. $\chi_w^e$, $\chi_w^b$ and $\chi_w^i$ are points in exterior, boundary and interior regions of the workspace, respectively.

### B. DS based Trajectory in the Workspace

The DS based obstacle avoidance approach in [16] is proposed to ensure the impenetrability of the obstacles, i.e. the DS is always in the exterior region of an obstacle. However, as described above, the DS in Fig.1(b) should never go out of the workspace.

At each point $\xi_w^b \in \chi_w^b$ on the inner surface of the workspace boundary in Fig.2, define $\tilde{\xi}_w^b = \xi - \xi_w^c$, we can get a tangential plane defined by its norm vector $n_w(\tilde{\xi}_w^b)$:

$$n_w(\tilde{\xi}_w^b) = \left[ -\frac{\partial \Gamma_w(\tilde{\xi}_w^b)}{\partial(\xi_w^b)_1} \quad -\frac{\partial \Gamma_w(\tilde{\xi}_w^b)}{\partial(\xi_w^b)_2} \quad -\frac{\partial \Gamma_w(\tilde{\xi}_w^b)}{\partial(\xi_w^b)_3} \right]^T \tag{6}$$

By extension, a deflection plane at each point $\xi \in \chi_w^i$ in the interior region of the workspace can be computed with normal vector:

$$n_w(\tilde{\xi}_w) = \left[ -\frac{\partial \Gamma_w(\tilde{\xi}_w)}{\partial(\xi)_1} \quad -\frac{\partial \Gamma_w(\tilde{\xi}_w)}{\partial(\xi)_2} \quad -\frac{\partial \Gamma_w(\tilde{\xi}_w)}{\partial(\xi)_3} \right]^T \tag{7}$$

A linear combination of a set of two linearly independent vectors that form a basis of the deflection plane, can describe every point on the deflection plane. Here, a set of vectors consists of $e_w^1(\tilde{\xi}_w)$ and $e_w^2(\tilde{\xi}_w)$ is chosen as:

$$\begin{cases} e_w^1(\tilde{\xi}_w) = \left[ (n_w(\tilde{\xi}_w))_2 \quad -(n_w(\tilde{\xi}_w))_1 \quad 0 \right]^T \\ e_w^2(\tilde{\xi}_w) = \left[ (n_w(\tilde{\xi}_w))_3 \quad 0 \quad -(n_w(\tilde{\xi}_w))_1 \right]^T \end{cases} \tag{8}$$

Similarly to the modulation matrix of a spherical obstacle determined in [16], the modulation matrix $M_w(\tilde{\xi}_w)$ of the workspace is given by:

$$M_w(\tilde{\xi}_w) = E_w(\tilde{\xi}_w) D_w(\tilde{\xi}_w) E_w(\tilde{\xi}_w)^{(-1)} \tag{9}$$

with a basis matrix $E_w(\tilde{\xi}_w)$ and an associated eigenvalue matrix $D_w(\tilde{\xi}_w)$ as:

$$E_w(\tilde{\xi}_w) = [n_w(\tilde{\xi}_w) \quad e_w^1(\tilde{\xi}_w) \quad e_w^2(\tilde{\xi}_w)] \quad (10)$$

$$D_w(\tilde{\xi}_w) = \mathrm{diag}(\lambda_w^1(\tilde{\xi}_w), \lambda_w^2(\tilde{\xi}_w), \lambda_w^3(\tilde{\xi}_w)) \quad (11)$$

where $\lambda_w^1(\tilde{\xi}_w) = \begin{cases} 1 - |\Gamma_w(\tilde{\xi}_w)| & |\Gamma_w(\tilde{\xi}_w)| > \lambda_w \\ 1 & |\Gamma_w(\tilde{\xi}_w)| \le \lambda_w \end{cases}$,

$\lambda_w^2(\tilde{\xi}_w) = \lambda_w^3(\tilde{\xi}_w) = \begin{cases} 1 + |\Gamma_w(\tilde{\xi}_w)| & |\Gamma_w(\tilde{\xi}_w)| > \lambda_w \\ 1 & |\Gamma_w(\tilde{\xi}_w)| \le \lambda_w \end{cases}$, $0 < \lambda_w < 1$.

Since $\Gamma_w(\tilde{\xi}_w)$ monotonically increases with $\|\tilde{\xi}_w\|$, when $\lambda_w = 0$, the matrices $D_w(\tilde{\xi}_w)$ and $M_w(\tilde{\xi}_w)$ converge to the identity matrix as the distance to the workspace center decreases. Therefore, the effect of the modulation matrix is maximum at the inner boundary of the workspace, and vanishes at the points near the workspace center.

In practice, the original DS should not be deformed by the modulation matrix except in the region near the inner workspace boundary. This can be achieved by choosing a proper value for $\lambda_w$ in Eq.(11). The effect of the modulation matrix disappears in more region of the workspace around the workspace center as $\lambda_w$ increases.

Similar to the obstacle avoidance method in [16], we can apply the workspace boundary modulation given by Eq.(9) to the original DS given by Eq.(1), then we have:

$$\dot{\xi} = M_w(\tilde{\xi}_w)f(\xi) \quad (12)$$

**Theorem 1** *Consider a three-dimensional convex workspace with boundary $\Gamma_w(\tilde{\xi}_w) = 1$ with respect to a reference point $\xi_w^c$ in the workspace. And the original DS given by Eq.(1) converges to a target point in the workspace. A trajectory $\{\xi\}_t$, that starts in the workspace, i.e. $\Gamma_w(\{\xi\}_0 - \xi_w^c) \le 1$, and evolves according to Eq.(12), will never go out of the workspace, i.e. $\Gamma_w(\{\xi\}_t - \xi_w^c) \le 1$, $t = 0..\infty$.* **Proof:** see Appendix A.

### C. Obstacle does not Intersect the Workspace Boundary

So far we have shown how the workspace modulation matrix $M_w(\tilde{\xi}_w)$ can be used to deform a DS such that it never go out of the workspace of a manipulator. In a case that there is an obstacle in the workspace and does not intersect the workspace boundary as shown in Fig.3(a), the original DS should be deformed by a proper dynamic modulation matrix such that it neither penetrate the obstacle nor go out of the workspace. This is a problem similar to the multi-obstacle avoidance problem described in [16].

When the obstacle does not intersect with the workspace boundary in Fig.3(a), as in [16], we can compute a dynamic modulation $M_o(\tilde{\xi}_o)$ of the obstacle as:

$$M_o(\tilde{\xi}_o) = E_o(\tilde{\xi}_o)D_o(\tilde{\xi}_o)E_o(\tilde{\xi}_o)^{(-1)} \quad (13)$$

with a basis matrix $E_o(\tilde{\xi}_o)$ and an associated eigenvalue matrix $D_o(\tilde{\xi}_o)$ as:

$$E_o(\tilde{\xi}_o) = [n_o(\tilde{\xi}_o) \quad e_o^1(\tilde{\xi}_o) \quad e_o^2(\tilde{\xi}_o)] \quad (14)$$

$$D_o(\tilde{\xi}_o) = \mathrm{diag}(\lambda_o^1(\tilde{\xi}_o), \lambda_o^2(\tilde{\xi}_o), \lambda_o^3(\tilde{\xi}_o)) \quad (15)$$

where $\lambda_o^1(\tilde{\xi}_o) = 1 - \frac{\omega_o(\tilde{\xi}_o)}{|\Gamma_o(\tilde{\xi}_o)|}$, $\lambda_o^2(\tilde{\xi}_o) = \lambda_o^3(\tilde{\xi}_o) = 1 + \frac{\omega_o(\tilde{\xi}_o)}{|\Gamma_o(\tilde{\xi}_o)|}$, and $\omega_o(\tilde{\xi}_o)$ is a weighting coefficient of the obstacle that can be computed according to [16]:

$$\omega_o(\tilde{\xi}_o) = \frac{(1 - \Gamma_w(\tilde{\xi}_w))}{(\Gamma_o(\tilde{\xi}_o) - 1) + (1 - \Gamma_w(\tilde{\xi}_w))} \quad (16)$$

and a corresponding weighting coefficient $\omega_w(\tilde{\xi}_w)$ of the workspace is computed as:

$$\omega_w(\tilde{\xi}_w) = \frac{(\Gamma_o(\tilde{\xi}_o) - 1)}{(\Gamma_o(\tilde{\xi}_o) - 1) + (1 - \Gamma_w(\tilde{\xi}_w))} \quad (17)$$

Then, the modulation matrix $M_w(\tilde{\xi}_w)$ of the workspace given by Eq.(9) can be modified as:

$$^wM_w(\tilde{\xi}_w) = E_w(\tilde{\xi}_w)^wD_w(\tilde{\xi}_w)E_w(\tilde{\xi}_w)^{(-1)} \quad (18)$$

with a modified eigenvalues matrix $^wD_w(\tilde{\xi}_w)$ as

$$^wD_w(\tilde{\xi}_w) = \mathrm{diag}(^w\lambda_w^1(\tilde{\xi}_w), \,^w\lambda_w^2(\tilde{\xi}_w), \,^w\lambda_w^3(\tilde{\xi}_w)) \quad (19)$$

where $^w\lambda_w^1(\tilde{\xi}_w) = \begin{cases} 1 - \omega_w(\tilde{\xi}_w)|\Gamma_w(\tilde{\xi}_w)| & |\Gamma_w(\tilde{\xi}_w)| > \lambda_w \\ 1 & |\Gamma_w(\tilde{\xi}_w)| \le \lambda_w \end{cases}$,

$^w\lambda_w^2(\tilde{\xi}_w) = ^w\lambda_w^3(\tilde{\xi}_w) = \begin{cases} 1 + \omega_w(\tilde{\xi}_w)|\Gamma_w(\tilde{\xi}_w)| & |\Gamma_w(\tilde{\xi}_w)| > \lambda_w \\ 1 & |\Gamma_w(\tilde{\xi}_w)| \le \lambda_w \end{cases}$.

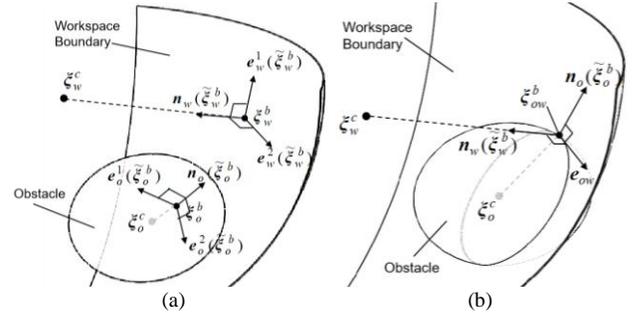

Figure 3. Illustration of the workspace boundary with a convex obstacle.

According to Eq.(16) and Eq.(17), we observe that $\omega_o(\tilde{\xi}_o) + \omega_w(\tilde{\xi}_w) = 1$, $0 \le \omega_o(\tilde{\xi}_o) \le 1$ and $0 \le \omega_w(\tilde{\xi}_w) \le 1$. At the obstacle boundary, we have $\omega_o(\tilde{\xi}_o) = 1$ and $\omega_w(\tilde{\xi}_w) = 0$, and vice versa. With the obstacle modulation matrix $M_o(\tilde{\xi}_o)$ given by Eq.(13) and the workspace modulation matrix $^wM_w(\tilde{\xi}_w)$ given by Eq.(18), we can compute a combined modulation matrix that consider the net effect of the obstacle and the workspace as:

$$\overline{M}(\tilde{\xi}) = M_o(\tilde{\xi}_o)^wM_w(\tilde{\xi}_w) \quad (20)$$

We can apply the combined modulation matrix $\overline{M}(\tilde{\xi})$ given by Eq.(20) to the original DS given by Eq.(1), then we have:

$$\dot{\xi} = \overline{M}(\tilde{\xi})f(\xi) \quad (21)$$

According to [16] and Theorem 1, a trajectory $\{\xi\}_t$, that starts in the workspace, and evolves according to Eq.(21), will neither go out of the workspace nor penetrate the obstacle.

## D. Obstacle Intersects the Workspace Boundary

In a case that an obstacle intersects the workspace boundary as shown in Fig.3(b), the modulation matrix $\overline{M}(\tilde{\xi})$ may lose efficacy, since $\Gamma_o(\tilde{\xi}_o) = \Gamma_w(\tilde{\xi}_w) = 1$ at the points on the intersection line between the obstacle and workspace boundary, then $\omega_o(\tilde{\xi}_o)$ and $\omega_w(\tilde{\xi}_w)$ are not numbers. Besides, this is a concave problem, and the original DS is always nonlinear, hence, current methods cannot deal with it [16, 17].

To solve the problem, consider a motion $\{\xi\}_t$, that starts in the workspace and outside the obstacle. When the motion $\{\xi\}_t$ does not reach any point on the intersection line, in order to ensure that the motion $\{\xi\}_t$ neither go out of the workspace nor penetrate the obstacle, according to [16] and Section C, the motion can evolve according to Eq.(21). When the motion $\{\xi\}_t$ reaches a point $\xi_{ow}^b$ on the intersection line between the obstacle and the workspace boundary, i.e. $\Gamma_w(\{\xi\}_t - \xi_w^c) = 1$ and $\Gamma_o(\{\xi\}_t - \xi_o^c) = 1$, define $\tilde{\xi}_w^b = \xi_{ow}^b - \xi_w^c$ and $\tilde{\xi}_o^b = \xi_{ow}^b - \xi_o^c$, we can compute the normal vector $n_w(\tilde{\xi}_w^b)$ of the workspace boundary and the normal vector $n_o(\tilde{\xi}_o^b)$ of the obstacle surface at the point $\xi_{ow}^b$ as follows:

$$n_w(\tilde{\xi}_w^b) = \left[ -\frac{\partial \Gamma_w(\tilde{\xi}_w^b)}{\partial(\xi)_1} \quad -\frac{\partial \Gamma_w(\tilde{\xi}_w^b)}{\partial(\xi)_2} \quad -\frac{\partial \Gamma_w(\tilde{\xi}_w^b)}{\partial(\xi)_3} \right]^T \quad (22)$$

$$n_o(\tilde{\xi}_o^b) = \left[ \frac{\partial \Gamma_o(\tilde{\xi}_o^b)}{\partial(\xi)_1} \quad \frac{\partial \Gamma_o(\tilde{\xi}_o^b)}{\partial(\xi)_2} \quad \frac{\partial \Gamma_o(\tilde{\xi}_o^b)}{\partial(\xi)_3} \right]^T \quad (23)$$

Then, a vector $e_{ow}(\xi_{ow}^b)$ that is perpendicular to the vector $n_w(\tilde{\xi}_w^b)$ and the vector $n_o(\tilde{\xi}_o^b)$ is given by:

$$e_{ow}(\xi_{ow}^b) = n_w(\tilde{\xi}_w^b) \times n_o(\tilde{\xi}_o^b) \quad (24)$$

Similarly to the modulation matrix given by Eq.(13), a modulation matrix $M_{ow}(\tilde{\xi}_o^b)$ of the point $\xi_{ow}^b$ on the intersection line between the obstacle and the workspace boundary is given by:

$$M_{ow}(\tilde{\xi}_o^b) = E_{ow}(\tilde{\xi}_o^b) D_{ow}(\tilde{\xi}_o^b) \text{pinv}(E_{ow}(\tilde{\xi}_o^b)) \quad (25)$$

with a basis matrix $E_{ow}(\tilde{\xi}_o^b)$ and an associated eigenvalue matrix $D_{ow}(\tilde{\xi}_o^b)$ as:

$$E_{ow}(\tilde{\xi}_o^b) = \left[ n_o(\tilde{\xi}_o^b) \quad e_{ow}(\xi_{ow}^b) \right] \quad (26)$$

$$D_{ow}(\tilde{\xi}_o^b) = \text{diag}(\lambda_o^1(\tilde{\xi}_o^b), \lambda_o^2(\tilde{\xi}_o^b)) \quad (27)$$

where $\lambda_o^1(\tilde{\xi}_o^b) = 1 - \frac{1}{|\Gamma_o(\tilde{\xi}_o^b)|}$, $\lambda_o^2(\tilde{\xi}_o^b) = 1 + \frac{1}{|\Gamma_o(\tilde{\xi}_o^b)|}$, $\text{pinv}(\cdot)$ is the pseudo inverse of $(\cdot)$.

We can apply the modulation matrix $M_{ow}(\tilde{\xi}_o^b)$ of the intersection line given by Eq.(25) to the original DS given by Eq.(1), then we have:

$$\dot{\xi} = M_{ow}(\tilde{\xi}_o^b) f(\xi) \quad (28)$$

**Theorem 2** *Consider a three-dimensional convex workspace with boundary $\Gamma_w(\tilde{\xi}_w) = 1$ with respect to a reference point $\xi_w^c$ in the workspace and a three-dimensional convex obstacle with boundary $\Gamma_o(\tilde{\xi}_o) = 1$ with respect to a reference point $\xi_o^c$ inside the obstacle. The obstacle intersects with the workspace boundary. A trajectory $\{\xi\}_t$, that starts in the workspace and outside the obstacle, i.e. $\Gamma_w(\{\xi\}_0 - \xi_w^c) \leq 1$ and $\Gamma_o(\{\xi\}_0 - \xi_o^c) \geq 1$, and evolves according to Eq.(21). When the trajectory $\{\xi\}_t$ reaches a point on the intersection line between the obstacle and the workspace boundary, i.e. $\Gamma_w(\{\xi\}_t - \xi_w^c) = 1$ and $\Gamma_o(\{\xi\}_t - \xi_o^c) = 1$, then the trajectory $\{\xi\}_t$ evolves according to Eq.(28) and vice versa, will neither go out of the workspace nor penetrate the obstacle, i.e. $\Gamma_w(\{\xi\}_t - \xi_w^c) \leq 1$ and $\Gamma_o(\{\xi\}_t - \xi_o^c) \geq 1$, $t = 0..\infty$.* **Proof:** see Appendix B.

Similarly to the safety margin given in [16], in practice, the condition, i.e. $\Gamma_w(\{\xi\}_t - \xi_w^c) = 1$ and $\Gamma_o(\{\xi\}_t - \xi_o^c) = 1$, given in Theorem 2 can be relaxed to $\beta_1 \leq \Gamma_w(\{\xi\}_t - \xi_w^c) \leq 1$ and $1 \leq \Gamma_o(\{\xi\}_t - \xi_o^c) \leq \beta_2$, where $0 < \beta_1 \leq 1$, $\beta_2 \geq 1$. $\beta_1$ and $\beta_2$ should be chosen as close to 1 as possible.

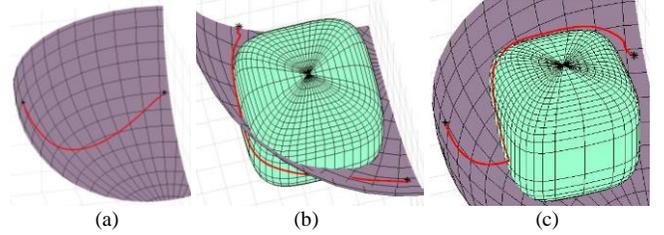

Figure 4. Illustration of the effectiveness of the proposed method.

Fig.4 illustrates the effectiveness of the proposed method. The red curve is DS based trajectory, the purple surface is the workspace boundary, the cyan cuboid is the obstacle, the cross mark is the start point of the trajectory and the star mark is the end point. Fig.4(a) shows an original DS given by Eq.(1), which is learned according to the method given in [18]. Fig.4(b) shows a DS deformed by the obstacle modulation matrix given in [16] that can avoid the obstacle. However, part of the trajectory is outside the workspace. Fig.4(c) shows a DS deformed by the proposed approach. We observe the DS neither penetrate the obstacle nor go out of the workspace, hence, the effectiveness of the approach is validated.

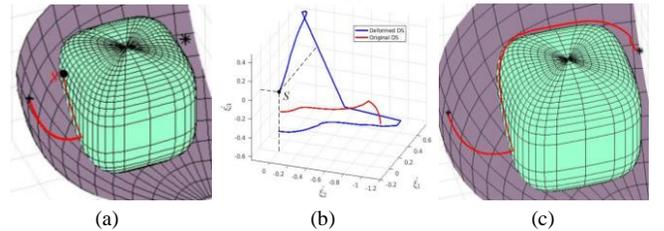

Figure 5. Illustration of a local minim point of the DS.

## E. Local Minima and Direction

According to [16], since the modulation matrix $M_{ow}(\tilde{\xi}_o^b)$ loses one rank, the modulated DS given by Eq.(28) may has some other equilibrium points besides the target point that could be local minima and/or saddle points, then, the DS will get stuck at those points [16]. To avoid the DS getting stuck into the local minima, when the norm of the velocity $\dot{\xi}$ given by Eq.(28) is less than a threshold value $v_{th}$, then, we replace $\dot{\xi}$ with $v_{th}\frac{\dot{\xi}}{\|\dot{\xi}\|}$. Fig.5(a) shows a local minima point $S$, and the corresponding velocities of deformed DS vanish at the point $S$ as shown in Fig.5(b).

According to Theorem 2, the velocity $\dot{\xi}$ given by Eq.(28) is parallel to the vector $\boldsymbol{e}_{ow}(\xi_{ow}^b)$ given by Eq.(24). However the direction of $\dot{\xi}$ is uncertain. We can set the direction of $\dot{\xi}$ to be the same as the direction of $\boldsymbol{e}_{ow}(\tilde{\xi}_{ow}^b)$ by replacing $\dot{\xi}$ with $\mathrm{sign}(\dot{\xi}^T \boldsymbol{e}_{ow}(\xi_{ow}^b))\dot{\xi}$. In addition, we can set the direction of $\dot{\xi}$ to be opposite to the direction of $\boldsymbol{e}_{ow}(\xi_{ow}^b)$ by replacing $\dot{\xi}$ with $-\mathrm{sign}(\dot{\xi}^T \boldsymbol{e}_{ow}(\xi_{ow}^b))\dot{\xi}$.

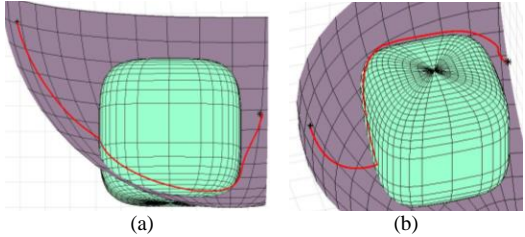

(a)          (b)

Figure 6. Illustration of two kinds of DS with different directions.

Fig.6(a) shows a DS with the direction of its velocity $\dot{\xi}$ given by Eq.(28) is set to be the same as the direction of $\boldsymbol{e}_{ow}(\xi_{ow}^b)$. Fig.6(b) shows a DS with the direction of its velocity $\dot{\xi}$ given by Eq.(28) is set to be the opposite to the direction of $\boldsymbol{e}_{ow}(\xi_{ow}^b)$. In both cases, the locations of the obstacle relative to the workspace are the same.

## IV. EXPERIMENT VALIDATION

To further verify the proposed method, we implement the proposed approach and the method presented in [16] on the 7-DOF arm of the UBTECH humanoid robot and carry out a set of comparative experiments. The arm as illustrated in Fig.7 is controlled at a rate of 100 Hz. The actual positions and the desired positions of the end-effector are converted to joint velocities using the task space controller given in [19] and the inverted kinematics of the arm. Since we don't have a visual detection system at present, we only consider stationary obstacles and assume that the positions and orientations of the obstacles are known. Since the obstacle avoidance methods can only ensure that a point will not penetrate the obstacle, a safety margin [16] is added to each obstacle.

Refer to Fig.4, in the experiments, an original trajectory is given for the end-effector. First, in the presence of the obstacle, i.e. a ball, the original trajectory is deformed by the modulation matrix of the ball in real-time according to the method in [16] to avoid the ball. However, because the method does not consider the workspace constraint of the arm, the deformed trajectory goes out of the workspace even though it avoids the ball. Hence, the end-effector will fail to track the trajectory and cannot reach the desired target.

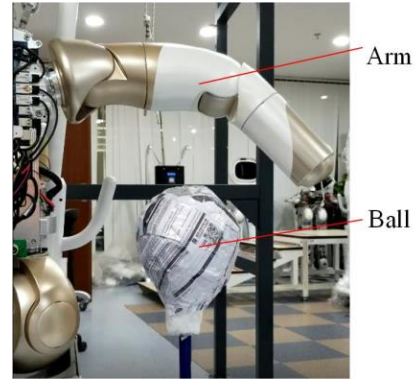

Figure 7. The experiment set-up.

Furthermore, in the presence of the ball, the original trajectory is deformed by the modulation matrix of the ball in real-time according to the proposed method to avoid the ball, while the end-effector tracks the deformed trajectory. The experiment results show that the end-effector tracking the deformed trajectory can avoid the ball and finally reach the desired target. In the experiment, the deformed trajectory neither penetrate the ball nor go out of the workspace. Therefore the experiment results indicate the effectiveness of the proposed approach.

According to [20], it is easy to apply the DS based obstacle avoidance approaches to the obstacle avoidance of moving obstacles and multiple obstacles. In the future, the extensions of the proposed method to moving obstacle avoidance and multi-obstacle avoidance will be presented. In addition, in practice, because the workspace boundaries are often more complex than spheres, we will further extend out the method for obstacle avoidance with complex workspace boundary constraints.

## V. CONCLUSION

In this paper, we presented a DS based obstacle avoidance approach for serial manipulators with limited workspace. The method works primarily by guiding the trajectory of the manipulator to evolve along an intersection line between the workspace boundary and the three-dimensional obstacle when the trajectory reaches a point on the intersection line. Besides, the trajectory can evolve along the intersection line in two opposite directions. We proved that the proposed approach can ensure that the trajectory neither penetrate the obstacle nor go out of the workspace without being stuck into local minima. The effectiveness of the approach was validated in several sets of comparative simulations. In addition, we implemented the method on the 7-DOF arm of the UBTECH humanoid robot. The experimental results show the effectiveness of the method. In the future, we will apply the approach to other problems such as concave obstacle avoidance and other robotic systems such as drones, underwater robots.

## VI. Appendix

### A. Proof of Theorem 1

To ensure that the trajectory $\{\xi\}_t$ never exceed the workspace boundary, the normal velocity at the boundary points $\xi_w^b \in \chi_w^b$ vanishes:

$$(\boldsymbol{n}_w(\tilde{\xi}_w^b))^T \dot{\xi}_w^b = 0 \quad (29)$$

With Eq.(9) and Eq.(12), Eq.(29) can be rewritten as:

$$(\boldsymbol{n}_w(\tilde{\xi}_w^b))^T \dot{\xi}_w^b = (\boldsymbol{n}_w(\tilde{\xi}_w^b))^T E_w(\tilde{\xi}_w^b) D_w(\tilde{\xi}_w^b) E_w(\tilde{\xi}_w^b)^{(-1)} \boldsymbol{f}(\cdot) \quad (30)$$

Since $\boldsymbol{n}_w(\tilde{\xi}_w^b)$ is perpendicular to the vectors $\boldsymbol{e}_w^1(\tilde{\xi}_w^b)$ and $\boldsymbol{e}_w^2(\tilde{\xi}_w^b)$, then Eq.(30) reduces to:

$$(\boldsymbol{n}_w(\tilde{\xi}_w^b))^T \dot{\xi}_w^b = \begin{bmatrix} a & 0 & 0 \end{bmatrix} D_w(\tilde{\xi}_w^b) E_w(\tilde{\xi}_w^b)^{(-1)} \boldsymbol{f}(\cdot) \quad (31)$$

where $a$ is a positive value. For each point on the workspace boundary, the eigenvalue $\lambda_w^1(\tilde{\xi}_w^b)$ is zero. Therefore, we get:

$$(\boldsymbol{n}_w(\tilde{\xi}_w^b))^T \dot{\xi}_w^b = \begin{bmatrix} 0 & 0 & 0 \end{bmatrix} E_w(\tilde{\xi}_w^b)^{(-1)} \boldsymbol{f}(\cdot) = 0 \quad (32)$$

### B. Proof of Theorem 2

To ensure that the trajectory $\{\xi\}_t$ neither go out of the workspace nor penetrate the obstacle, when the trajectory $\{\xi\}_t$ does not reach any point on the intersection line between the obstacle and the workspace boundary, the aforementioned performance is ensured according to [16] and Theorem 1.

When the trajectory $\{\xi\}_t$ reaches a point on the intersection line between the obstacle and the workspace boundary, the normal velocity of the obstacle boundary and the normal velocity of the workspace boundary at the points $\xi_{ow}^b \in \chi_w^b \cap \chi_o^b$ on the intersection line vanish:

$$(\boldsymbol{n}_o(\tilde{\xi}_o^b))^T \dot{\xi}_{ow}^b = 0 \quad (33)$$

$$(\boldsymbol{n}_w(\tilde{\xi}_w^b))^T \dot{\xi}_{ow}^b = 0 \quad (34)$$

With Eq.(25) and Eq.(28), Eq.(33) and Eq.(34) can be rewritten as:

$$(\boldsymbol{n}_o(\tilde{\xi}_o^b))^T \dot{\xi}_{ow}^b = (\boldsymbol{n}_o(\tilde{\xi}_o^b))^T E_{ow}(\tilde{\xi}_o^b) D_{ow}(\tilde{\xi}_o^b) \text{pinv}(E_{ow}(\tilde{\xi}_o^b)) \boldsymbol{f}(\cdot) \quad (35)$$

$$(\boldsymbol{n}_w(\tilde{\xi}_w^b))^T \dot{\xi}_{ow}^b = (\boldsymbol{n}_w(\tilde{\xi}_w^b))^T E_{ow}(\tilde{\xi}_o^b) D_{ow}(\tilde{\xi}_o^b) \text{pinv}(E_{ow}(\tilde{\xi}_o^b)) \boldsymbol{f}(\cdot) \quad (36)$$

Since $\boldsymbol{e}_{ow}(\xi_{ow}^b)$ is perpendicular to the vectors $\boldsymbol{n}_o(\tilde{\xi}_o^b)$ and $\boldsymbol{n}_w(\tilde{\xi}_w^b)$, then Eq.(35) and Eq.(36) reduce to:

$$(\boldsymbol{n}_o(\tilde{\xi}_o^b))^T \dot{\xi}_{ow}^b = \begin{bmatrix} a_o & 0 \end{bmatrix} D_{ow}(\tilde{\xi}_o^b) \text{pinv}(E_{ow}(\tilde{\xi}_o^b)) \boldsymbol{f}(\cdot) \quad (37)$$

$$(\boldsymbol{n}_w(\tilde{\xi}_w^b))^T \dot{\xi}_{ow}^b = \begin{bmatrix} a_w & 0 \end{bmatrix} D_{ow}(\tilde{\xi}_o^b) \text{pinv}(E_{ow}(\tilde{\xi}_o^b)) \boldsymbol{f}(\cdot) \quad (38)$$

where $a_o$ and $a_w$ are real values. For each point on the intersection line, the eigenvalue $\lambda_o^1(\tilde{\xi}_o^b)$ is zero. Therefore, we get:

$$(\boldsymbol{n}_o(\tilde{\xi}_o^b))^T \dot{\xi}_{ow}^b = \begin{bmatrix} 0 & 0 \end{bmatrix} \text{pinv}(E_{ow}(\tilde{\xi}_o^b)) \boldsymbol{f}(\cdot) = 0 \quad (39)$$

$$(\boldsymbol{n}_w(\tilde{\xi}_w^b))^T \dot{\xi}_{ow}^b = \begin{bmatrix} 0 & 0 \end{bmatrix} \text{pinv}(E_{ow}(\tilde{\xi}_o^b)) \boldsymbol{f}(\cdot) = 0 \quad (40)$$

As stated above, according to [16] and Theorem 1, with Eq.(39) and Eq.(40), Theorem 2 is proved.